\documentclass[runningheads]{llncs}
\usepackage[final]{eccv}
\usepackage{eccvabbrv}
\usepackage{graphicx}
\usepackage{booktabs}
\usepackage[accsupp]{axessibility}  % Improves PDF readability for those with disabilities.

\usepackage{bbm}
\usepackage{booktabs}
\usepackage{multirow}
\usepackage{bigstrut}
\usepackage{makecell}

\newcommand{\cred}[1]{\textbf{\color{red}#1}}

\usepackage[table]{colortbl}
\usepackage{wrapfig}
\usepackage[pagebackref,breaklinks,colorlinks,citecolor=eccvblue]{hyperref}
\usepackage{hyperref}
\usepackage{orcidlink}

\begin{document}
\title{SparseSSP: 3D Subcellular Structure Prediction from Sparse-View Transmitted Light Images}
\titlerunning{SparseSSP: Sparse-View Subcellular Structure Prediction}

\author{Jintu Zheng\inst{1,2}\and
    Yi Ding\inst{3}\and
    Qizhe Liu\inst{1}\and
    Yi Cao\inst{3}\and
    Ying Hu\inst{1}\and
    Zenan Wang\inst{1}
}

\authorrunning{J. Zheng et al.}
\institute{
Shenzhen Institute of Advanced Technology, Chinese Academy of Sciences, China\\
\email{\{jt.zheng,zn.wang1\footnote[4]{\scriptsize Corresponding author. This work was accepted to ECCV2024. Project page: \url{https://github.com/JintuZheng/SparseSSP}, and the code would be available after August 1st, 2024.},ying.hu\}@siat.ac.cn}
\and
University of Chinese Academy of Sciences, Beijing, China\\
\and
Jinan University, Shandong, China\\
\email{dingyi@stu.ujn.edu.cn,ise\_caoy@ujn.edu.cn}
}

\maketitle
\begin{abstract}
    Traditional fluorescence staining is phototoxic to live cells, slow, and expensive;
    thus, the subcellular structure prediction (SSP) from transmitted light (TL) images is emerging as a label-free, faster, low-cost alternative.
    However, existing approaches utilize 3D networks for one-to-one voxel level dense prediction, which necessitates a frequent and time-consuming Z-axis imaging process.
    Moreover, 3D convolutions inevitably lead to significant computation and GPU memory overhead.
    Therefore, we propose an efficient framework, SparseSSP, predicting fluorescent intensities within the target voxel grid in an efficient paradigm instead of relying entirely on 3D topologies.
    In particular, SparseSSP makes two pivotal improvements to prior works.
    First, SparseSSP introduces a one-to-many voxel mapping paradigm, which permits the sparse TL slices to reconstruct the subcellular structure.
    Secondly, we propose a hybrid dimensions topology, which folds the Z-axis information into channel features, enabling the 2D network layers to tackle SSP under low computational cost.
    We conduct extensive experiments to validate the effectiveness and advantages of SparseSSP on diverse sparse imaging ratios, and our approach achieves a leading performance compared to pure 3D topologies.
    SparseSSP reduces imaging frequencies compared to previous dense-view SSP (i.e., the number of imaging is reduced up to 87.5\% at most), which is significant in visualizing rapid biological dynamics on low-cost devices and samples.
  \keywords{Subcellular Structure Prediction \and Hybrid Dimensions Topology \and Microscopy Image}
\end{abstract}

\begin{figure*}[htttp]
    \centering
    \includegraphics[width=\linewidth]{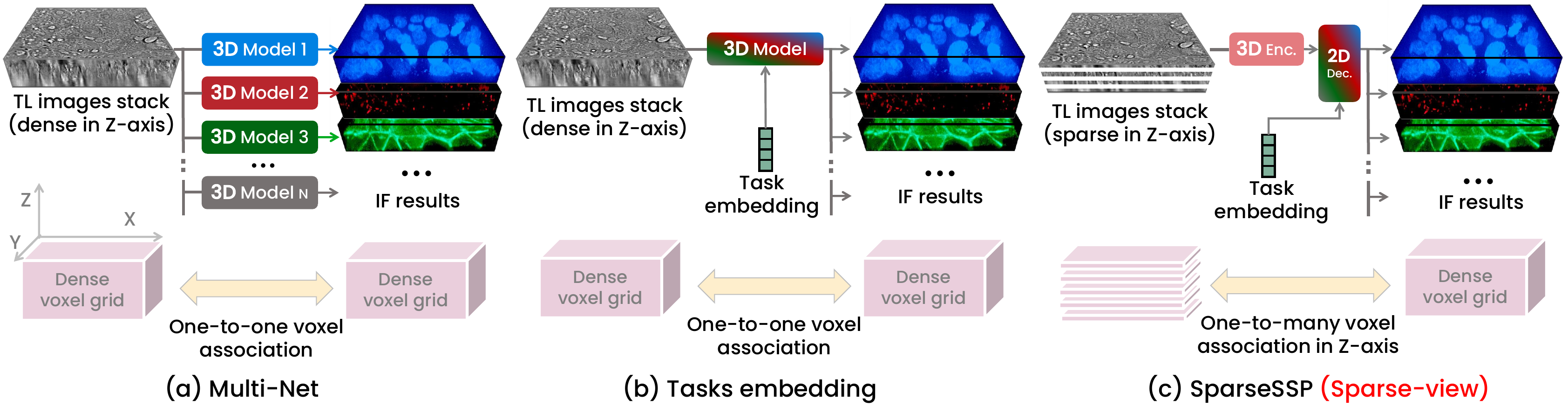}
    \caption{\textbf{Comparisons on SSP implementations.} (a) Multi-Net adopts a set of independent 3D models for SSP. (b) Task embedding is an improved strategy which can learn all tasks relied on the one-hot encoding, however, this is still require a dense imaging procedure. Prolonged imaging time and high computational cost are unfriendly. (c) SparseSSP implements an efficient hybrid dimensions topology which only requires less TL imaging slices to achieve the excellent performance.}
    \label{fig:banner1}
\end{figure*}

\section{Introduction}
\label{sec:intro}
% what is the Fl?
Understanding information at the subcellular level is pivotal for biological and medical studies  \cite{thul2017subcellular, christopher2021subcellular, gut2018multiplexed, carlton2020membrane, wolff2020molecular, guo2018visualizing, bock2020mitochondria}.
Fluorescence microscopy revolutionized modern biology by permitting labeled imaging and quantifying subcellular structures of interest.
This technology employs antibodies conjugated with various fluorescent dyes to cells, enabling the visualization of rich structural information in 3D through distinct fluorescent signals (i.e., immunofluorescence images).

% what is the SSP? & Desen imaging is time-cosuming.
Fluorescence staining requires expensive and advanced instrumentation and time-consuming preparation of materials \cite{challen1,challen2,im2019introduction}.
Moreover, significant phototoxicity and photobleaching also damage the live cells \cite{icha2017phototoxicity}.
An emerging technology, namely Subcellular Structure Prediction (SSP) \cite{repmode,fnet}, enables direct prediction of 3D immunofluorescence (IF) from transmitted light (TL) images via 3D vision networks, significantly reduces sample preparation costs and is low phototoxic to live cells \cite{wolff2020molecular}.
SSP predicts the fluorescent intensities scalar in a predefined 3D voxel grid, i.e., predicting an occupancy value for each voxel, which can be seen as a dense regression problem.
Existing SSP approaches require the dense imaging process (i.e., a motor is required to drive the lens to scan layer by layer on the z-axis) for better data quality, which would cost a prolonged imaging time (e.g., in AllenCell collection \cite{fnet}, each subcellular type is imaged for up to 2.5 hours on a Zeiss spinning disk microscope).
Prolonged imaging time is unfriendly to capturing the biodynamic process; the physiological motion, such as cell respiration, introduces the scanning position offset.
Additionally, the nutrients produce wakes and motion artifacts, making observing live cell kinematics accurately challenging.
Therefore, developing an approach for sparse-view SSP is desired, which makes it significant in visualizing rapid biological dynamics on low-cost devices and samples.

% 3D model is large and high cost.
The essence of the SSP is learning the distribution of the fluorescence signal within the target voxel grid according to the subcellular structure characteristics in TL images.
A logical approach is to train a one-to-one voxel mapping between the input and output 3D spaces via pure 3D topologies.
FNet \cite{fnet} employs multiple 3D models to learn diverse subcellular structures, as shown in \cref{fig:banner1}(a).
RepMode \cite{repmode} introduces a task embedding solution for SSP, enabling multiple tasks to be embedded into a single model, reducing computational costs significantly, as shown in \cref{fig:banner1}(b).
Pure 3D modules (e.g., 3D convolutions), owing to their overhead on GPU memory and computational cost, pose substantial challenges in biological imaging applications.

% what is our solution? why it works?
Motivated by the aforementioned observations, we propose an efficient framework, SparseSSP, for predicting fluorescent intensities on less TL images (i.e., sparse-view) with a hybrid dimensions paradigm, as shown in \cref{fig:banner1}(c).
SparseSSP mainly makes two pivotal improvements to prior SSP works.
First, SparseSSP introduces a one-to-many voxel mapping paradigm, enabling more minor microscopy imaging frequencies to capture the subcellular structure.
We interpolate the sparse-view space into the pseudo voxel grid and aim to learn an implicit upsample optimization on the Z-axis.
SparseSSP embeds the Z-axis information into channel features, enabling efficient 2D network layers to address this 3D prediction task.
Recently, FlashOcc \cite{flashocc} proposed a pure 2D manner for 3D occupancy prediction \cite{Occ1,Occ2}, which validates that the 2D networks are feasible to learn stereo representation in absolute 3D space.
In addition, channel rearrangement \cite{shi2016real} in super-resolution demonstrates the potentiality of reconstructing information missing in spatial dimensions via a channel-to-spatial manner.
Inspired by the above works, we implement a Z-axis spatial information to channel cross-transformation, aiming to model the Z-axis sparse-view via efficient 2D decoding layers.
We conduct extensive experiments to explore diverse hybrid dimensions topologies for sparse-view SSP. This substantiates that the 2D decoding layers are superior in modeling the sparse-view voxel grid after depth-to-channel transformation, demonstrating more excellent performance than pure 3D topologies.

Our contributions can be summarized as follows:

\begin{itemize}
    \item We propose an efficient hybrid dimensions topology for SSP, enabling the 2D network layers to tackle the 3D voxel level prediction, which simultaneously trades off computational cost and precision, demonstrating its potential for more biological researches.
    \item We conduct extensive experiments to validate the effectiveness and advantages of SparseSSP on diverse sparse imaging ratios.
    \item To the best of the author's knowledge, we are the pioneers to study SSP from sparse-view TL images. Our approach significantly reduces the number of required imaging frequencies by up to 87.5\% while maintaining leading performance compared to pure 3D topologies.
\end{itemize}

\section{Related Work}
\label{sec:rwork}
\subsection{Partially Labeling Issue in SSP}
Diverse subcellular structures are annotated in different images because of the high laborious cost \cite{repmode} and limitations in simultaneous multiple staining.
It is a partial labeling issue similar to many medically dense prediction tasks, e.g., multi-organ segmentation. Fortunately, many mature solutions \cite{fnet,DoDNet,TSNs,Tgnet,repmode,CondNet} are available for multi-task learning.

\noindent
\textbf{Multi-Net.}
Multi-Net adopts independent networks to learn tasks \cite{fnet, jo2021label, cheng2021single, kandel2020phase}.
FNet \cite{fnet} is the typical approach to employ multiple 3D UNets to predict diverse subcellular structures, which is inefficient and inflexible in tuning the network parameters.

\noindent
\textbf{Task Embedding.}
Some approaches \cite{CondNet,TSNs,DoDNet,Tgnet} attempt to encode each task as the one-hot embedding feature in the network inference.
Dmitriev et al. \cite{CondNet} propose a conditional framework CondNet on multi-task learning, while Zhang et al. \cite{DoDNet} propose the DoDNet with the task controller and a dynamic head to formulate the multi-task problem as a single class task.
Tgnet \cite{Tgnet} improves DoDNet \cite{DoDNet}, embedding more task-guiding functional blocks in the network. These multi-head solutions share the feature extractor and employ different task embedding approaches to achieve varied output.
Meanwhile, Zhou et al. \cite{repmode} propose the RepMode to learn gating re-parameterize to generate specialized parameters for each subcellular structure, achieving state-of-the-art performance on dense-view TL images. However, all task embedding methods are focused on the dense-view SSP. Furthermore, especially for 3D prediction, these models heavily rely on computationally expensive pure 3D convolutions, which demand significant GPU memory and computational resources.

\subsection{Sparse-View Techniques}
Sparse-view techniques have emerged as a prominent research area in biological and medical imaging \cite{han2018framing,hu2020hybrid,lee2018deep,zhang2018sparse}, garnering significant attention and interest. The sparse-view impact significantly brings up more physical gain than just the software advantages. For example, sparse-view techniques can reduce radiation dose in CT reconstruction with fewer projection times \cite{li2023learning}.
Similarly, less imaging time in SSP also reduces the phototoxicity of live cells \cite{fnet}.
More importantly, reducing imaging times to capture the subcellular structures allows biologists to observe the rapid biological dynamics in a low-cost preparation to understand subcellular-level activities better.

\section{SparseSSP}
\label{sec:med}
\subsection{Problem Definition}
\noindent
\textbf{View Definition.}
Let $\mathbf{S}$ denote the target 3D voxel grid of size $D_s \times H \times W$, and we would like to obtain the subcellular structures in $\mathbf{S}$.
The microscope requires TL imaging layer by layer in the Z-axis to reconstruct the structure information in this voxel grid.
Let $\mathcal{I}$ denote imaging 3D voxel grid of size $D_i \times H \times W$.
Assume the most minor interval of micromotor is the same as the voxel Z-axis resolution limitation.
In a dense view, the motor must walk at least $D_i$ steps to cover the voxel grid $\mathbf{S}$, i.e., $D_s = D_i$; this guarantees maximum imaging quality. However, this would slow down the imaging dramatically because the frequent acceleration and deceleration movements in a very short Z-axis interval require high fineness of the micromotor.
To meet such high-precision imaging requirements, micromotors are usually mechanically designed to vary slowly to ensure accuracy.
The goal of sparse-view SSP is to predict the voxel grid from $\mathbf{I} \to \mathbf{S}$ under the condition $D_s < D_i$ , which can reduce the imaging frequencies.
In order to make it more suitable for the actual situation of the imaging scenes, let $\mathbf{r}$ as the sparsity ratio, which is satisfying that the $D_i$ is the multiple of $\mathbf{r}$ with $D_s$ (which can be easily controlled by the micromotor walking interval), i.e., $\mathbf{r} \times D_i = D_s$, especially when $\mathbf{r}=1$ it is the dense-view.

\begin{figure*}[htttp]
    \centering
    \setlength{\belowcaptionskip}{-0.5cm}
    \includegraphics[width=\linewidth]{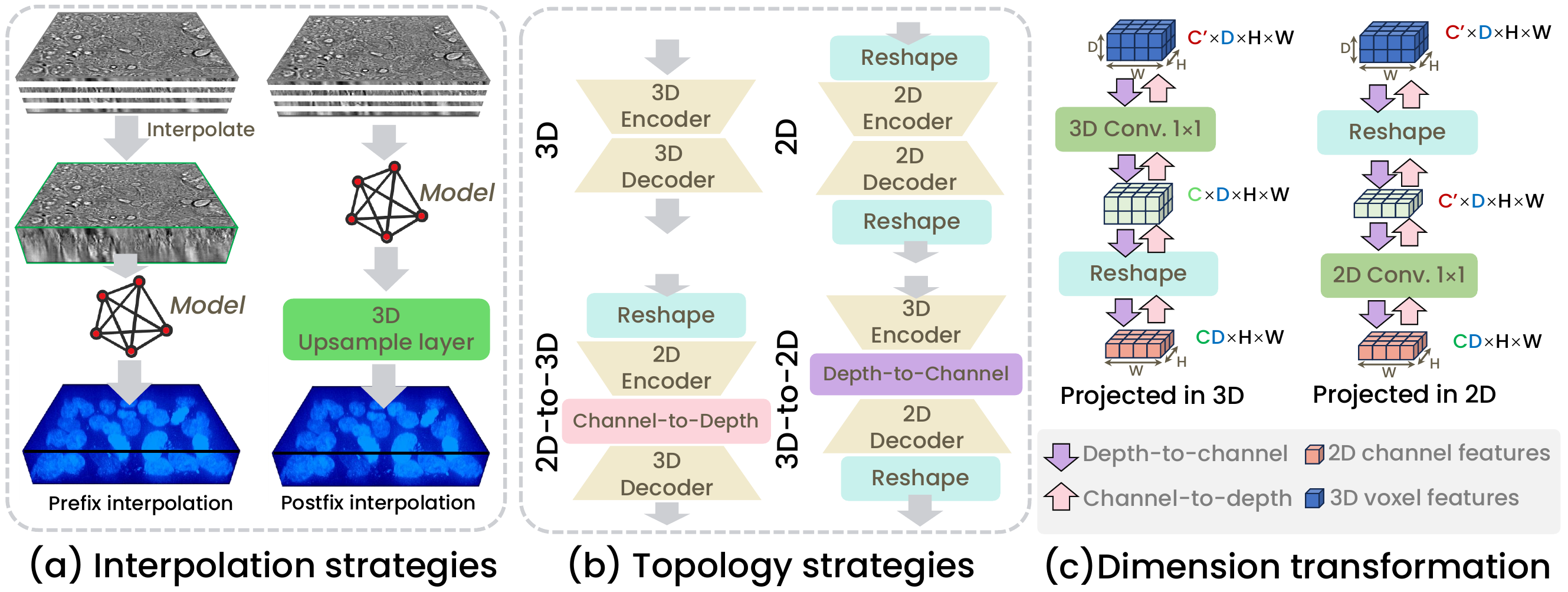}
    \caption{\textbf{Optional strategies in SparseSSP.} (a) demonstrates two strategies for one-to-many mapping in the Z-axis. (b) shows four combinations for different dimension encoders and decoders. (c) demonstrates diverse approaches to implementing the dimension transformation.}
    \label{fig:options}
\end{figure*}

\noindent
\textbf{Data Definition.}
Following prior works \cite{fnet,repmode}, we assume that each TL input contains only one fluorescent label.
Let $\mathcal{P}=\{(\mathbf{x}_n,\mathbf{y}_n, l_n)\}_{n=1}^{N}$ denotes the SSP dataset in the sparse-view setting with $N$ samples.
The $n$-th image $\mathbf{x}_n \in \mathcal{P}_n$ is associated with the label $\mathbf{y}_n \in \mathcal{P}_n$,
where $\mathbf{y}_n = \mathbbm{R}^{D_n \times H_n \times W_n}$ denotes the fluorescent label, and the $\mathbf{x}_n = \mathbbm{R}^{(D_n/\mathbf{r}) \times H_n \times W_n}$ denotes the TL input ($D_n/\mathbf{r}$ images).
The $\mathbf{y}_n$ satisfies the view definition $\mathbf{S}$, and the $\mathbf{x}_n$ satisfies the view definition $\mathbf{I}$.
We define the $T$ as the number of tasks, i.e., the number of subcellular structures that require learning, and the $l_n \in \mathcal{L} = \{1, 2, ..., T\}$ represents that $\mathbf{y}_n$ is the label of the $l_n$-th subcellular structure.

\subsection{One-to-Many Mapping in Z-axis}
Sparse-view is sparse on the Z-axis which means that the Z-axis information is incomplete in describing the target voxel grid $S$.
In contrast to dense-view, which solely focuses on prediction, sparse-view necessitates tackling both prediction and upsampling within a network.
Here, we propose two strategies, prefix and postfix interpolation, as shown in \cref{fig:options}(a).
The prefix strategy involves performing interpolation prior to the network, with the aim of mapping into a pseudo voxel grid.
The network requires to learn a mapping, $M_{pre}: \mathbf{S'} \to \mathbf{S}$, where $\mathbf{S'}$ denotes this pseudo voxel grid of size $D_s \times H \times W$.
The postfix strategy is firstly to learn a one-to-one voxel grid mapping, $M_{post}: \mathbf{I} \to \mathbf{I'} \to \mathbf{S}$, then interpolate it to $S$, where $\mathbf{I'}$ denotes the network output voxel grid of size $D_i \times H \times W$.
The prefix interpolation is the traditional algorithm implementation, e.g., nearest or trilinear, while the postfix interpolation is a learnable module that adopts deconvolution as the upsampling layer.
The prefix strategy generates the pseudo voxel grid before the model input; in this strategy, the Z-axis information is implicitly restored through learning the fluorescence prediction.
In contrast, the postfix strategy, learning the restore procedure through an explicit upsampling layer, separates two processing purposes.
We argue that implicitly learning the spatial context restoration benefits the subcellular prediction because the prefix strategy provides more prior structure knowledge with interpolation.

\subsection{Hybrid Dimensions Topology}
All prior works employ pure 3D topologies for prediction because the input and output is the one-to-one voxel associated to 3D space.
Despite their accuracy, convolutions or deconvolutions implemented in the 3D manner are cost more GPU memory and FLOPs than 2D.
So it is natural to reduce the 3D operations.
To this end, we propose a hybrid dimensions topology eschewing the use of costing 3D convolutions in encoding and decoding.
In \cref{fig:options}(b), we demonstrate four topology strategies (i.e., 2D, 3D, 2-to-3D and 3-to-2D).
Here we adopt the native UNet as the baseline (i.e., contains 5 layers in encoder and decoder), and the UNet layers follow the residual-style convolution architecture.

\noindent
\textbf{Dimension transformation.}
This core implementation of SparseSSP enables the 3-to-2D or 2-to-3D feature transformation. Transforming to a 2D feature permits all existing advanced modules or technologies achieved on 2D features to migrate into our framework.
It is an encouraging improvement, enabling more extensive development based on SparseSSP in the future.
As depicted in \cref{fig:options}(c), there are two transformation routes, i.e., depth-to-channel and channel-to-depth. We introduce the route of depth-to-channel first.
In simple terms, we compress the 3D feature into a 2D space, and it looks like we are folding the dimensions from the Z-axis into channels. Here, we discuss two projection implementations, as shown in the right part of \cref{fig:options}(c).
The embedding in 2D means the feature is projected in 2D space after the channel arrangement.
Embedding in 3D means that the feature is projected in 3D space (as shown in the left part of \cref{fig:options}(c)), which keeps relatively complete 3D structural representation information before the channel arrangement.
Dimension transformation via channel arrangement would break the structural expression of 3D space; thus, it is more rational to embed in 3D, though it costs a little more computation than embedding in 2D.
The route of channel-to-depth is similar to the above implementations, and the most significant difference is to unsqueeze the 2D feature to a 3D representation.

\begin{figure*}[htttp]
    \centering
    \setlength{\belowcaptionskip}{-0.5cm}
    \includegraphics[width=\linewidth]{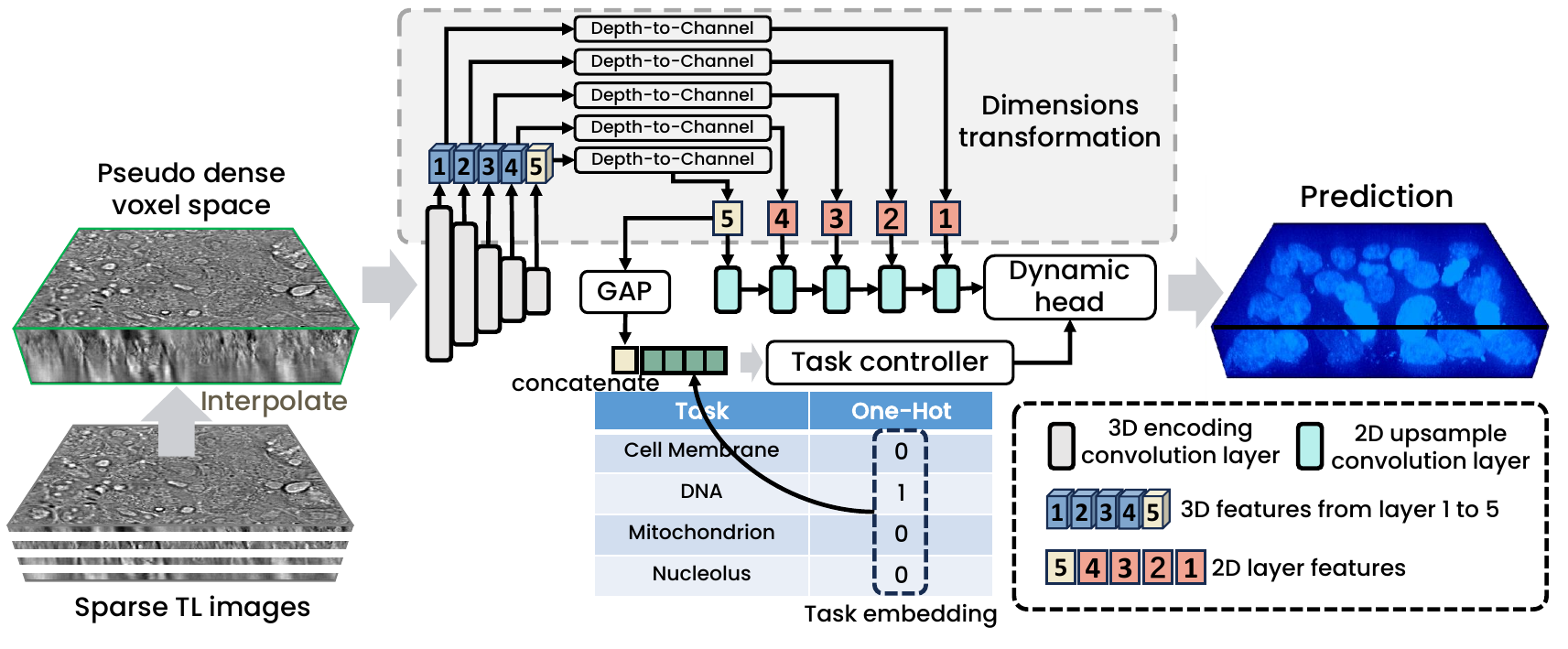}
    \caption{\textbf{Overview of SparseSSP with 3-to-2D topology and prefix interpolation.} It is an example following the DoDNet \cite{DoDNet} style that utilizes the task controller and dynamic head. The framework's task embedding approach can be conveniently changed to other technologies.}
    \label{fig:ooview}
\end{figure*}

\noindent
\textbf{Pure 2D Topology.}
This 2D network for 3D SSP is utilized to compare with the following hybrid dimensional topologies.
Inspired by FlashOcc's \cite{flashocc} fully 2D operations for 3D occupancy prediction, we would like to verify the effectiveness of this implementation in SSP.
We first reshape the input voxel into 2D space.
Specifically, the Z-axis dimension is squeezed to the channel dimension, as formulated like $\mathbf{x}_n \in \mathbbm{R}^{1 \times D_n \times H_n \times W_n}$  transforms to $\mathbf{x'}_n \in \mathbbm{R}^{D_n \times H_n \times W_n}$, where $\mathbf{x}_n$ denotes the TL input.
For the network output, we reshape the prediction from $\mathbf{y'}_n \in \mathbbm{R}^{D_n \times H_n \times W_n}$ to $\mathbf{y}_n \in \mathbbm{R}^{1 \times D_n \times H_n \times W_n}$.
In this topology, the depth $D_n$ of the 3D feature is treated as the channel dimension during the encoding and decoding.

\noindent
\textbf{Hybrid of 3-to-2D.}
The hybrid of 3-to-2D topology contains a 3D encoder and a 2D decoder.
After the 3D feature extraction, a dimension transformation in a depth-to-channel manner is required to convert the 3D features into 2D space.
There are 5 layers in the encoder, following the RepMode layer settings, let $\mathbf{C}_e=\{32, 64, 128, 256, 256\}$  denotes as the output channels of 3D space.
Note that the last layer is the bottleneck layer.
We perform depth-to-channel transformation for each layer.
Let $k\in [1,5]$ denote the layer index, and the resolution of 3D features can be formatted as $x_k \in \mathbbm{R}^{C_k \times D_k \times H_k \times W_k}, C_k \in \mathbf{C}_e$.
Let $U$ denote the upsample embedding uniform dimension (we fixed the setting as 256 in the experiments; see more comparisons about this hyper-parameter in supplementary materials), which can trade off the model's performance and scale.
Here, we discuss two projection implementations in formulation.

1) If we assume that the projection is embedded in 3D space.
Define the $\hat{x_k} = \Gamma(\lambda, \mu, x_k)$ as the projection function, where the $\lambda$ and $\mu$ are the input and output channels, where $\hat{x_k}$ denotes the projected feature; here, it must satisfy that $\lambda = C_k$ and $\mu = C_u/D_k$ (e.g., in the output of bottleneck layer, $U = 256, C_k = 256, D_k = 2, H_k = 8, W_k = 8$, and the $\lambda$ and $\mu$ must be defined as 256 and 128).
We employ a 3D convolution on kernel size of 1 as the projecting function.
Then reshape the $\hat{x_k} \in \mathbbm{R}^{\mu \times D_k \times H_k \times W_k}$ to $\mathbbm{R}^{\mu D_k \times H_k \times W_k}$.

2) If we assume that the projection is embedded in 2D space.
We firstly reshape $\hat{x_k} \in \mathbbm{R}^{C_k \times D_k \times H_k \times W_k}$ to $\mathbbm{R}^{C_k D_k \times H_k \times W_k}$.
For the projection function $\Gamma$, the parameters $\lambda$ and $\mu$ are required to be formulated as $\lambda = C_k D_k$ and $\mu = U$.
We employ a 2D convolution as the projecting function.

After the dimension transformation, we employ the 2D decoding layers to upsample these features.
The output channel of the final convolution head equals the length of the Z-axis in the target voxel grid, e.g., an inference patch is $32\times 128 \times 128$, and the final output channel is 32.

\noindent
\textbf{Hybrid of 2-to-3D.}
The hybrid 3-to-2D topology contains a 2D encoder and a 3D decoder.
We first reshaped the 3D TL input into a 2D shape like the pure 2D topology. Different from 3-to-2D, we employ the channel-to-depth transformation, as shown in \cref{fig:options}(b) and (c).
The 2D encoder can be changed to other backbones, and the decoder is the same as the typical 3D topology.

\subsection{Task Eembeding Mechanisms}
SparseSSP is designed as a plug-and-play paradigm, meaning all components can be conveniently changed to existing state-of-the-art technologies. For example, the task embedding approach, e.g., task controller (DoDNet) \cite{DoDNet}, task guiding enhancement (Tgnet) \cite{Tgnet}, task switching (TSNs) \cite{TSNs}, and reparameterize diverse expert (RepMode) \cite{repmode,liu20233d}.
We demonstrate an example of a task controller and dynamic head following DoDNet in \cref{fig:ooview}.

\section{Experiments}
\label{sec:exp}
\subsection{Experimental setup}

\noindent
\textbf{Benchmark.}
We conduct the experiments based on AllenCell collection \cite{fnet}, containing 12 subcellular structures (i.e., ActinFilament, ActomBundle, CellMembrane, Desmosome, DNA, EndopReticulum, GolgiApparatus, Microtubule, Mitochondria, NuclearEnvelope, Nucleolus and TightJunction).
We follow the data distribution (i.e., for each structure, select 25\% of samples for evaluation and 10\% of the rest for the test) and preprocessing (e.g., z-scored normalization for voxels and data augmentation) as identical to the RepMode \cite{repmode} (we directly adopt the prefabricated data splitting from the RepMode \cite{repmode}).
The fluorescence labels are resized to make the voxel correspond to $0.29 \times 0.29 \times 0.29 \mu m^3$ , while the TL input images are resized to make the voxel correspond to $(0.29 \mathbf{r}) \times 0.29 \times 0.29 \mu m^3$ ($\mathbf{r}$ is the sparsity rate).
We mainly compare the sparsity ratios of 2, 4, and 8, and the sparsity ratio 8 means that it can already reduce 87.5\% imaging frequencies compared to the source microscopy session.

% \subsubsection{Evaluation Metrics.}
\noindent
\textbf{Evaluation Metrics.}
Following the RepMode, we adopt three metrics for SSP, i.e., the mean absolute error (MAE), mean square error (MSE), and the coefficient of determination ($R^2$) to measure the quality of prediction.
MSE and MAE are in arbitrary ranges, and the smaller value is better, while the $R^2$ ranges from 0 to 1, and the larger value is better performance.
Moreover, unless otherwise specified, we compute the mean value from 12 datasets as the overall performance.

\begin{table*}[t]
    \centering
    \scriptsize
    \setlength{\tabcolsep}{1.6pt}
    \setlength{\belowcaptionskip}{-0.2cm}
    \renewcommand{\arraystretch}{1.}
    \begin{tabular}{l|cc|ccc|ccc|ccc}
        \hline
                                  & \multicolumn{2}{c|}{} & \multicolumn{3}{c|}{$\mathbf{r}=2$} & \multicolumn{3}{c|}{$\mathbf{r}=4$} & \multicolumn{3}{c}{$\mathbf{r}=8$}                                                                                                          \\
        \hline
        Approach                  & Interp.               & Topology                            & MSE                                 & MAE                                & $R^2$        & MSE          & MAE          & $R^2$        & MSE          & MAE          & $R^2$        \\
        \hline
        RepMode                   & None                  & 2D                                  & .6055                               & .4420                              & .3632        & .6096        & .4513        & .3542        & .6109        & .4523        & .3493        \\
        RepMode                   & None                  & 3-to-2D                             & .5668                               & .4376                              & .4032        & .5769        & .4376        & .3813        & .6041        & .4423        & .3690        \\
        DoDNet                    & None                  & 2D                                  & .6103                               & .4509                              & .3524        & .6238        & .4623        & .3243        & .6298        & .4602        & .3103        \\
        DoDNet                    & None                  & 3-to-2D                             & .5781                               & .4368                              & .3813        & .5860        & .4421        & .3794        & .6013        & .4469        & .3638        \\
        \hline
        RepMode \cite{repmode}    & post                  & 3D                                  & .5455                               & .4234                              & .4193        & .5539        & .4360        & .4137        & .5886        & .4424        & .3736        \\
        \rowcolor{gray!20}RepMode & pre                   & 3D                                  & .5210                               & .4248                              & .4453        & .5320        & .4197        & .4359        & .5687        & .4359        & .3984        \\
        RepMode                   & post                  & 2D                                  & .6043                               & .4471                              & .3613        & .6154        & .4500        & .3460        & .6189        & .4562        & .3324        \\
        \rowcolor{gray!20}RepMode & pre                   & 2D                                  & .5704                               & .4532                              & .3984        & .5812        & .4383        & .3857        & .5896        & .4484        & .3849        \\
        RepMode                   & post                  & 2-to-3D                             & .5458                               & .4236                              & .4250        & .5523        & .4312        & .4153        & .5896        & .4447        & .3790        \\
        \rowcolor{gray!20}RepMode & pre                   & 2-to-3D                             & .5135                               & .4173                              & .4543        & .5299        & .4194        & .4397        & .5624        & .4286        & .3971        \\
        RepMode                   & post                  & 3-to-2D                             & .5234                               & .4132                              & .4589        & .5356        & .4232        & .4264        & .5780        & .4313        & .3862        \\
        \rowcolor{gray!20}RepMode & pre                   & 3-to-2D                             & \textbf{.5069}                        & .4140                              & \textbf{.4616} & \textbf{.5159} & .4136        & \textbf{.4523} & .5468        & .4222        & .4207        \\
        \hline
        DoDNet \cite{DoDNet}      & post                  & 3D                                  & .5343                               & .4241                              & .4243        & .5541        & .4313        & .4117        & .5768        & .4381        & .3861        \\
        \rowcolor{gray!20}DoDNet  & pre                   & 3D                                  & .5173                               & .4257                              & .4512        & .5392        & .4318        & .4288        & .5572        & .4316        & .4103        \\
        DoDNet                    & post                  & 2D                                  & .6012                               & .4487                              & .3643        & .6045        & .4412        & .3632        & .6123        & .4561        & .3520        \\
        \rowcolor{gray!20}DoDNet  & pre                   & 2D                                  & .5751                               & .4372                              & .3883        & .5793        & .4342        & .3734        & .5823        & .4413        & .3699        \\
        DoDNet                    & post                  & 2-to-3D                             & .5486                               & .4329                              & .4232        & .5554        & .4382        & .4143        & .5774        & .4367        & .3903        \\
        \rowcolor{gray!20}DoDNet  & pre                   & 2-to-3D                             & .5244                               & .4185                              & .4463        & .5367        & .4150        & .4221        & .5535        & .4234        & .4145        \\
        DoDNet                    & post                  & 3-to-2D                             & .5354                               & .4213                              & .4201        & .5475        & .4335        & .4172        & .5634        & .4318        & .4032        \\
        \rowcolor{gray!20}DoDNet  & pre                   & 3-to-2D                             & .5128                               & \textbf{.4118}                       & .4516        & .5229        & \textbf{.4133} & .4452        & \textbf{.5440} & \textbf{.4209} & \textbf{.4236} \\
        \hline
    \end{tabular}
    \caption{\textbf{Impact of the interpolation strategies.} The prefix strategies have a clear lead compared with no interpolation and the postfix. The $\mathbf{r}$ is the sparsity ratio of the input voxel grid. No interpolation can only be implemented in a 2D decoding manner. Gray-back areas are the better settings.}
    \label{tab:interp_cmp}
    % \vspace{-5mm}
\end{table*}

% \subsubsection{Training Settings.}
\noindent
\textbf{Training Settings.}
For 3D topologies, we use 1000 training epochs, the Adam optimizer, MSELoss function, batch size of 8, and evaluate every 20 epochs, following RepMode.
For 3D and 2D blending networks, we extend training to 3000 epochs. Pure 2D topologies undergo 7000 epochs of training with consistent settings. Despite needing more epochs to converge, the hybrid dimensions topology trains faster than pure 3D layers.
Detailed resource costs are in \cref{sec:compute}. Using the Gaussian sliding inference strategy from RepMode \cite{repmode}, the window size is $32\times128\times128$. 
All experiments were conducted on an NVIDIA RTX 3090 with 24GB memory, and our experiments are easily replicable on devices with at least 18GB of memory.

\subsection{Ablation studies}
We conduct ablative experiments to evaluate each proposal in our framework.
We implement the SOTA multi-task methods (i.e., DoDNet \cite{DoDNet}, Tgnet \cite{Tgnet}, TSNs \cite{TSNs}, CondNet \cite{CondNet}, and RepMode \cite{repmode}) on our topologies strategies (i.e.,3D, 2D, 2-to-3D and 3-to-2D) with different interpolation strategies.
We employ the 3D channel projection for hybrid dimensions topologies in both channel-to-depth and depth-to-channel unless stated otherwise.

\begin{table*}[t]
    \scriptsize
    \centering
    \setlength{\belowcaptionskip}{-0.2cm}
    \setlength{\tabcolsep}{1.6pt}
    \renewcommand{\arraystretch}{1.}
    \begin{tabular}{l|c|ccc|ccc|ccc}
  \hline
 & \multicolumn{1}{c|}{} & \multicolumn{3}{c|}{$\mathbf{r}=2$} & \multicolumn{3}{c|}{$\mathbf{r}=4$} & \multicolumn{3}{c}{$\mathbf{r}=8$}   \\
  \hline
  Proj. space& Topology  & MSE     & MAE     & $R^2$  & MSE   & MAE   & $R^2$ & MSE   & MAE   & $R^2$ \\
  \hline
  \rowcolor{gray!20}3D & 3-to-2D   & .5128   & .4118   & .4516  & .5229 & .4133 & .4452 & .5440 & .4209 & .4236 \\
  2D                   & 3-to-2D   & .5351   & .4149   & .4336  & .5405 & .4221 & .4143 & .5660 & .4389 & .3969 \\
  \hline
  \rowcolor{gray!20}3D & 2-to-3D   & .5244 & .4185 & .4463& .5367  & .4150  & .4221  & .5535  & .4234    & .4145    \\
  2D                   & 2-to-3D   & .5313 & .4142 & .4377& .5437  & .4265  & .4246  & .5718  & .4382   &  .3913   \\
  \hline
    \end{tabular}
    \caption{\textbf{Impact of the projection space.}
    We compare the different implementations in dimension transformation based on the DoDNet \cite{DoDNet}. The projection in 3D space performed better than the projection in 2D space. It can maintain more spatial context for 3-to-2D topologies, while for 2-to-3D, it can pre-reconstruct a 3D space and refine it during the layer-by-layer upsampling. 
    }
    \label{tab:embed_appro}
    % \vspace{-5mm}
\end{table*}

\begin{table*}[htttp]
    \centering
    \scriptsize
    \setlength{\tabcolsep}{2.4pt}
    \setlength{\belowcaptionskip}{-0.2cm}
    \renewcommand{\arraystretch}{1.}
    \begin{tabular}{l|c|ccc|ccc|ccc}
        \hline
          & \multicolumn{1}{c|}{} & \multicolumn{3}{c|}{$\mathbf{r}=2$} & \multicolumn{3}{c|}{$\mathbf{r}=4$} & \multicolumn{3}{c}{$\mathbf{r}=8$}                   \\
        \hline
        Approach                  & Topology              & MSE         & MAE         & $R^2$      & MSE          & MAE          & $R^2$        & MSE          & MAE          & $R^2$        \\
        \hline
        TSNs \cite{TSNs}          & 3D                    & .5540       & .4325       & .4135      & .5663        & .4358        & .4009        & .5871        & .4463        & .3796        \\
        TSNs                      & 2D                    & .6024       & .4474       & .3640      & .6046        & .4488        & .3617        & .6111          & .4513          & .3551          \\
        TSNs                      & 2-to-3D               & .5343       & .4198       & .4323      & .5620        & .4322        & .4025        & .5846        & .4396        & .3897          \\
        \rowcolor{gray!20}TSNs    & 3-to-2D               & .5164       & .4101       & .4519      & .5328        & .4174        & .4351        & .5542        & .4262        & .4132        \\
        \hline
        CondNet \cite{CondNet}    & 3D                    & .5388       & .4279       & .4225      & .5425        & .4334        & .4113        & .5633        & .4446        & .3928        \\
        CondNet                   & 2D                    & .5864       & .4419       & .3821      & .5878        & .4517        & .3814        & .5943        & .4528          & .3683          \\
        CondNet                   & 2-to-3D               & .5267       & .4134       & .4416      & .5413        & .4221        & .4094        & .5612        & .4148        & .4001          \\
        \rowcolor{gray!10}CondNet & 3-to-2D               & .5211       & .4123       & .4501      & .5326        & .4195        & .4281        & .5498        & .4202        & .4205        \\
        \hline
        DoDNet \cite{DoDNet}      & 3D                    & .5173       & .4257       & .4512      & .5392        & .4318        & .4288        & .5572        & .4316        & .4103        \\
        DoDNet                    & 2D                    & .5751       & .4372       & .3883      & .5793        & .4342        & .3734        & .5823        & .4413         & .3699          \\
        DoDNet                    & 2-to-3D               & .5244       & .4185       & .4463      & .5367        & .4150        & .4221        & .5535        & .4234          & .4145          \\
        \rowcolor{gray!20}DoDNet  & 3-to-2D               & .5128       & .4118       & .4516      & .5229        & \cred{.4133} & .4452        & .5440        & \cred{.4209} & .4236        \\
        \hline
        \hline
        RepMode \cite{repmode}    & 3D                    & .5210       & .4248       & .4453      & .5320        & .4197        & .4359        & .5687        & .4359        & .3984        \\
        RepMode                   & 2D                    & .5704       & .4532       & .3984      & .5812        & .4383        & .3857          & .5896        & .4484        & .3849          \\
        RepMode                   & 2-to-3D               & .5135       & .4173       & .4543      & .5299        & .4194        & .4397        & .5624        & .4286        & .3971          \\
        \rowcolor{gray!20}RepMode & 3-to-2D               & .5069       & .4140       & .4616      & .5159        & .4136        & .4523        & .5468        & .4222        & .4207        \\
        \hline
        Tgnet \cite{Tgnet}        & 3D                    & .5224       & .4259       & .4459      & .5421        & .4315        & .4257        & .5691        & .4437        & .3981        \\
        Tgnet                     & 2D                    & .5690       & .4327       & .4034      & .5713        & .4368        & .3947        & .5773        & .4351        & .3896        \\
        Tgnet                     & 2-to-3D               & .5034       & .4122       & .4519      & .5231        & .4143        & .4503        & .5498        & .4291        & .4164        \\
        \rowcolor{gray!20}Tgnet   & 3-to-2D               & \cred{.4995}& \cred{.4113}& \cred{.4693}& \cred{.5158} & .4166        & \cred{.4526} & \cred{.5386} & .4231        & \cred{.4292} \\
        \hline
    \end{tabular}
    \caption{
        \textbf{Topologies comparisons for sparse-view SSP.}
        We compare the performance of existing mainstream multi-task approaches implemented in diverse topologies. For hybrid dimensions topologies, i.e., the 3-to-2D and 2-to-3D, employ the 3D space projection in the dimension transformation. All of these experiments are prefix interpolations. The red-bold row has less MSE and MAE and the largest $R^2$, and the rows in gray-back are the winning strategy with a group. 
    }
    \label{tab:topo_cmp}
    % \vspace{-10mm}
\end{table*}

% \subsubsection{Interpolation Strategies.}
\noindent
\textbf{Interpolation Strategies.}
We implement the active interpolation in SparseSSP for one-to-many voxel mapping in Z-axis.
We compare the performance on two approaches, i.e., prefix and postfix, and the details are provided in \cref{tab:interp_cmp}(row5-20).
For a fairness, we utilize the nearest algorithm in all interpolations.
In addition, there is a naive alternative to interpolation, but only for the topologies with the 2D decoding manner.
That is, directly define the output channels as same as the Z-axis length of voxel grid (e.g., the Z-length of target grid is 32, and set the output channels of final head as 32).
We demonstrate the performance on this alternative in \cref{tab:interp_cmp}(row1-4), which are described as "None".
The results indicate that, when employing the prefix interpolation technique, both DoDNet and RepMode representational models outperform the postfix and none interpolation.
Prefix interpolation enables the incorporation of prior spatial knowledge into the model learning process, thereby enhancing the performance of sparse-view SSP; therefore, we perform the prefix strategy on the following experiments.

% \subsubsection{Impact of the Projection Operation.}
\noindent
\textbf{Projection Operation.}
Projection is an essential operation in hybrid dimensions topology, and the space projected in which dimension means maintaining the spatial context in where.
We compare two implementations in \cref{tab:embed_appro}, i.e., projected in 3D or 2D, on DoDNet, and it can be observed that projection in 3D outperforms 2D space.
Projection in 3D maintains more spatial context for 3-to-2D topologies, while for 2-to-3D, it can pre-reconstruct a 3D space and refine it during the layer-by-layer upsampling. Maintaining a relatively complete feature transformation is crucial for hybrid dimensions topology.

\begin{figure*}[htttp]
    \centering
    \setlength{\belowcaptionskip}{-0.3cm}
    \includegraphics[width=\linewidth]{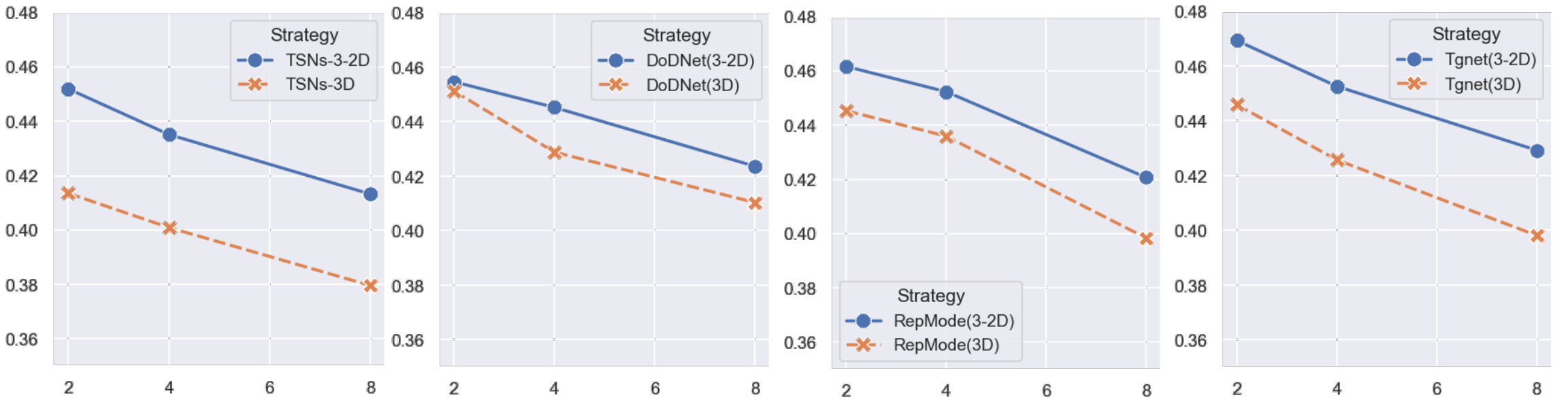}
    \caption{\textbf{Trend of $R^2$ value as sparsity ratio increased from 2 to 8.} Hybrid dimensions topology 3-to-2D (i.e., the blue lines in the figure) shows a slower decay and higher global scores than pure 3D topology (i.e., the orange lines). 
    }
    \label{fig:trend}
    % \vspace{-3mm}
\end{figure*}

% \subsubsection{Topology Strategies on Diverse Methodologies.}
\noindent
\textbf{Topology Strategies on Diverse Methodologies.}
We compare 5 SOTA multi-task methodologies, and the performance of diverse topology strategies are as listed in \cref{tab:topo_cmp}.
We demonstrate a trend of $R^2$ values as the sparsity ratio increased from 2 to 8 in \cref{fig:trend}, and it can be obverse that the quality of the prediction keeps getting worse as the sparsity ratio increases; however, the 3-to-2D topology shows a slower decay than pure 3D networks.
From \cref{tab:topo_cmp}, it can be observed that the hybrids, i.e., 3-to-2D and 2-to-3D, outperform pure 2D and 3D on diverse sparsity ratios, and the pure 2D strategies are much poorer than other implementations.
Notably, the 3-to-2D strategy based on Tgnet \cite{Tgnet} achieves the best score in all metrics, and RepMode \cite{repmode}, which is the SOTA method in dense-view SSP, fell to Tgnet by a narrow margin.
Hybrid dimensions topologies have the leading advantages on sparse-view SSP, especially the 3-to-2D strategy, which encodes in 3D space and decodes the 2D features. It means that the 3D feature extractor is essential for a hybrid strategy.
Meanwhile, the Z-axis dimension is folded into the channel features after the transformation of dimensions, and it permits the sparse Z-axis information to be refined during the 2D layer-by-layer upsample process.
In contrast, 2D features from 2D extractors are more difficult for downstream 3D decoders to understand, and the cases involved in 2D encoding would lead to a longer convergence process. Therefore, we will discuss more potential factors in the following sections.

\begin{wraptable}{tr}{0.38\textwidth}
    \scriptsize
    \setlength{\belowcaptionskip}{-0.2cm}
    \centering
        \setlength{\tabcolsep}{1.3pt}
        \renewcommand{\arraystretch}{1.}
        \begin{tabular}{l|c|ccc}
            \hline
                   & \multicolumn{1}{c|}{} & \multicolumn{3}{c}{$\mathbf{r}=2$}                 \\
            \hline
            epochs & Topology              & MSE                                & MAE   & $R^2$ \\
            \hline
            7000   & 2D                    & .5751                              & .4372 & .3883 \\
            14000  & 2D                    & .5623                              & .4252 & .4176 \\
            \hline
            3000   & 2-to-3D               & .5244                              & .4185 & .4463 \\
            7000   & 2-to-3D               & .5231                              & .4183 & .4485 \\
            \hline
            \hline
        \end{tabular}
    \caption{We increase the training epochs on DoDNet \cite{DoDNet} to study whether the 2D encoder requires a longer convergence time.}
    \label{tab:training_epochs}
    % \vspace{-6mm}
\end{wraptable}

% \subsubsection{Impact of Training epochs.}
\noindent
\textbf{More Training Epochs.}
We save the best checkpoint on validation data during training. For 3D and 3-to-2D topologies, increasing training epochs to 7000 does not produce a better checkpoint; however, epochs increasing benefits the topologies with 2D encoders. As shown in \cref{tab:training_epochs}, more training epochs would lead to a little better performance, which means that there is still room for optimization. 

\begin{figure*}[htttp]
    \centering
    \setlength{\belowcaptionskip}{-0.5cm}
    \includegraphics[width=\linewidth]{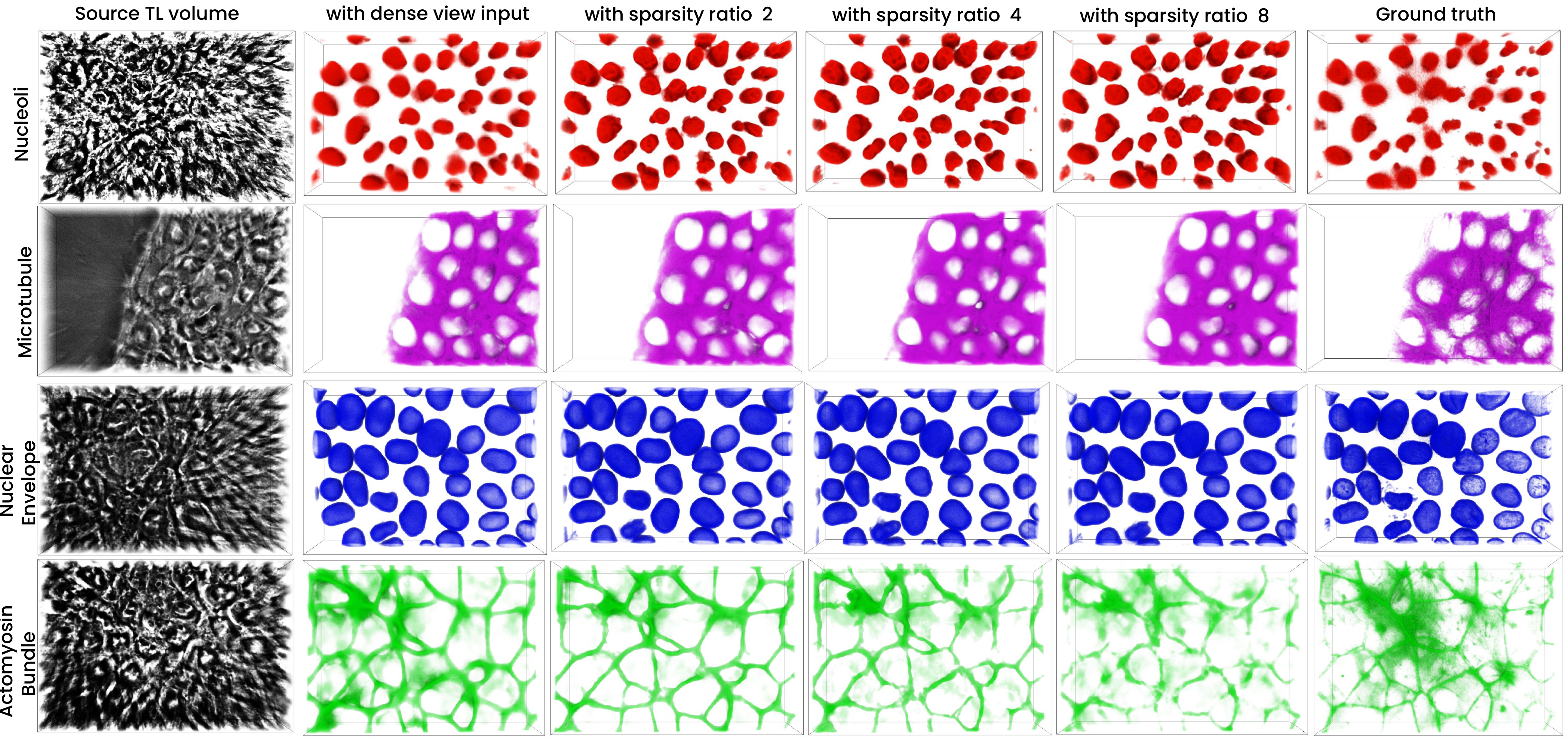}
    \caption{\textbf{Visualization of subcellular structure prediction.} A stereoscopic volume display from a top-down Z-axis view. We compare results of four structures using dense-view and sparse-view TL images at different sparsity ratios. The second column shows results from RepMode (best for dense-view SSP), and columns three to six show results from Tgnet with the best 3-to-2D topology for sparse-view.}
    \label{fig:visz}
    % \vspace{-3mm}
\end{figure*}

\begin{table*}[b]
    \scriptsize
    \centering
    \setlength{\belowcaptionskip}{-0.2cm}
    \setlength{\tabcolsep}{1.6pt}
    \renewcommand{\arraystretch}{1.}
    \begin{tabular}{l|c|ccc|cccccc}
        \hline
        & \multicolumn{1}{c|}{} & \multicolumn{3}{c|}{$\mathbf{r}=2$} & \multicolumn{3}{c}{$\mathbf{r}=8$}                                                                     \\
        \hline
        Backbone    & Topology              & MSE                                 & MAE                                & $R^2$          & MSE            & MAE            & $R^2$          \\
        \hline
        RestNet18   & 2D                    & .6024                               & .4474                              & .3640          & .6111          & .4513          & .3551          \\
        Swin-Base   & 2D                    & .5867                               & .4412                              & .3800          & .5977          & .4490          & .3687          \\
        ViT-Base    & 2D                    & .5889                               & .4420                              & .3812          & .5813          & .4323          & .3846          \\
        DINOv2-Base & 2D                    & .6143                               & .4563                              & .3507          & .6324          & .4682          & .3340          \\
        ViT-Large   & 2D                    & .5674                               & .4425                              & .3995          & .5713          & .4402          & .3893          \\
        \hline
        RestNet18   & 2-to-3D               & .5343                               & .4198                              & .4323          & .5846          & .4396          & .3897          \\
        Swin-Base   & 2-to-3D               & .5266                               & .4132                              & .4501          & .5623          & .4323          & .4133          \\
        ViT-Base    & 2-to-3D               & .5300                               & .4157                              & .4385          & .5641          & .4428          & .3970          \\
        DINOv2-Base & 2-to-3D               & .5543                               & .4305                              & .4124          & .5952          & .4561          & .3628          \\
        ViT-Large   & 2-to-3D               & \textbf{.5155}                      & \textbf{.4098}                     & \textbf{.4559} & \textbf{.5524} & \textbf{.4240} & \textbf{.4194} \\
        \hline

    \end{tabular}
    \caption{\textbf{Ablations on 2D backbones, based on the TSNs \cite{TSNs}}, generally exhibit poorer performance than other implementations; thus, we replaced the 2D backbone for it. It is encouraging to see that the results are improved with the transformer backbones, which means that there is potential to migrate more powerful 2D network technologies into our framework, e.g., some of the latest transformer designs.}
    \label{tab:2d_backbone}
    % \vspace{-5mm}
\end{table*}

% \subsubsection{Impact of 2D Backbone in 2-to-3D Topology.}
\noindent
\textbf{Impact of 2D Backbone in 2-to-3D Topology.}
The 2D feature extractor involved topologies, i.e., the pure 2D and 2-to-3D, which enables the replacement of the 2D modules for its components. We found that increasing the epochs would improve its results, which means it is possible to increase the volume of 2D network parameters.
We conducted an ablation study on TSNs, which demonstrated inferior performance compared to other implementations, to assess the impact of backbone replacement. We employ a SOTA transformer-based backbone Swin-Transformer in its base architecture (Swin-B) for replacement. Swin-B is much larger than the native backbone ResNet18 of TSNs on parameters.
In \cref{tab:2d_backbone}, we can observe a better performance after increasing the parameters, which means there is currently a correlation between the topologies' performance in 2D feature extractors and the design of their encoders. It shows the potential to introduce more powerful technologies from 2D network design, e.g., some of the latest transformer designs. 

\begin{figure*}[ttp]
    \centering
    \includegraphics[width=\linewidth]{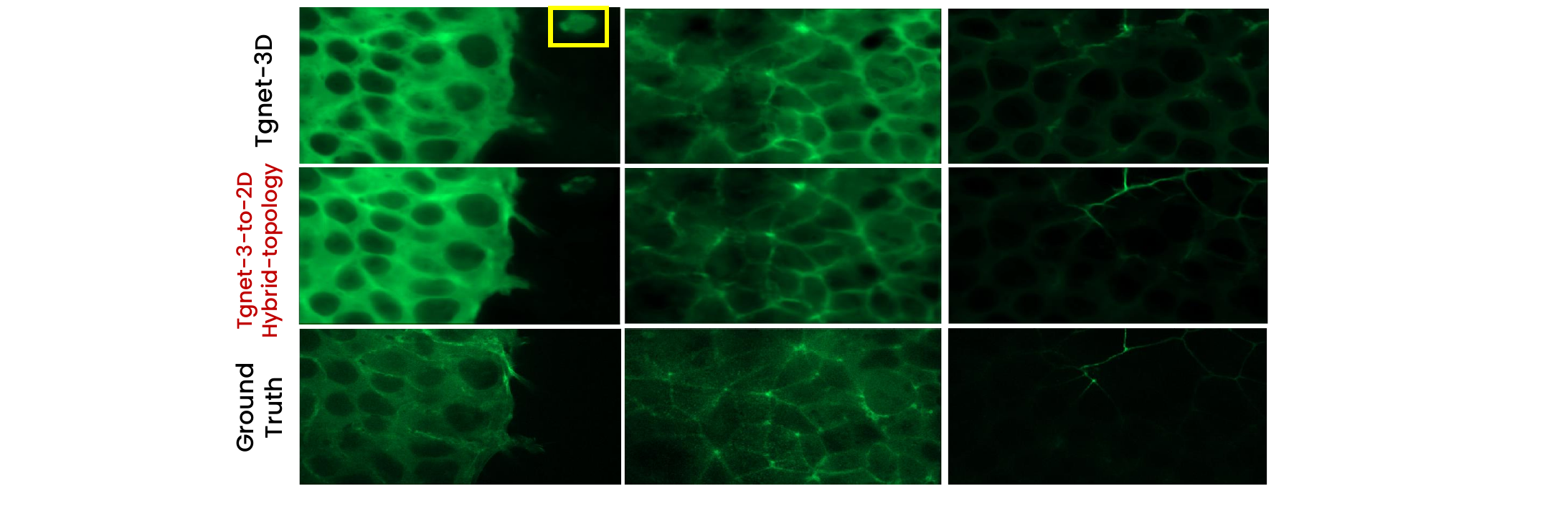}
    \caption{Comparison of pure 3D and our hybrid topology soulution on TGNet. The yellow bounding boxes are the regions with significant prediction errors.}
    \label{fig:visz1}
\end{figure*}

% \subsubsection{Summarized Analysis.}
\noindent
\textbf{Summarized for Ablations.}
We demonstrate the performances of two interpolation strategies and four topologies' combinations in sparse-view SSP.
We found that the combination of prefix interpolation and 3-to-2D strategies demonstrated significantly better performance than others (a visual comparison under the premise of sparsity rate 2 as shown in \cref{fig:visz1}).
% We found that the combination of prefix interpolation and 3-to-2D strategies demonstrated significantly better performance than others.
Moreover, the projection in 3D space during the dimension transformation of hybrid dimensions topologies is better than in 2D space.
In addition, training epochs and backbone types would also affect the topologies of 2D and 2-to-3D. The effectiveness and advantages of our proposals in this paper are validated in the above ablation studies.

\begin{table*}[tp]
    \scriptsize
    % \footnotesize
    \centering
    \setlength{\tabcolsep}{0.9pt}
    \setlength{\belowcaptionskip}{-0.2cm}
    \renewcommand{\arraystretch}{1.0}
    \begin{tabular}{l|c|cc|cc|cc}
        \hline
                 &          & \multicolumn{2}{c|}{GPU Infer.} & \multicolumn{2}{c|}{GPU Train.} & \multicolumn{1}{c}{Computation}                       \\
        \hline
        Approach & Topology & time(s/iter)                    & Mem.(MiB)                       & Speed(iter/s)                   & Mem.(MiB) & MACs    \\
        \hline
        RepMode  & 3D       & 4.47                            & 9122                            & 0.89                            & 17843     & 66.29G  \\
        RepMode  & 2D       & 0.66                            & 2472                            & 3.31                            & 3548      & 2.33G   \\
        RepMode  & 2-to-3D  & 1.86                            & 3666                            & 1.86                            & 8984      & 30.11G  \\
        RepMode  & 3-to-2D  & 2.55                            & 5428                            & 1.29                            & 15521     & 43.47G  \\
        \hline
        DoDNet   & 3D       & 2.11                            & 4710                            & 2.64                            & 16054     & 113.86G \\
        DoDNet   & 2D       & 0.35                            & 1982                            & 4.70                            & 2692      & 1.82G   \\
        DoDNet   & 2-to-3D  & 1.04                            & 2897                            & 3.31                            & 4384       & 41.02G  \\
        DoDNet   & 3-to-2D  & 1.35                            & 4124                            & 4.53                            & 14362     & 76.61G  \\
        \hline
        TSNs     & 3D       & 1.09                            & 6862                            & 7.19                            & 10458     & 55.70G  \\
        TSNs     & 2D       & 0.37                            & 1720                            & 31.12                           & 2793      & 2.05G   \\
        TSNs     & 2-to-3D  & 0.86                            & 2028                            & 10.2                            & 5729      & 13.94G   \\
        TSNs     & 3-to-2D  & 0.91                            & 3554                            & 8.91                            & 7522      & 46.77G  \\
        \hline
    \end{tabular}
    \caption{
        \textbf{Comparisons on resource consumption}. We compare the TSNs \cite{TSNs}, DoDNet \cite{DoDNet} and RepMode \cite{repmode} in diverse topologies.
        Hybrid dimensions topologies demonstrate less resource consumption than pure 3D, especially in MACs. The number of iterations in training is the number of the loss backward operations. 
    }
    \label{tab:cost}
    % \vspace{-10mm}
\end{table*}

\begin{table*}[t]
    \centering
    \setlength{\belowcaptionskip}{-0.2cm}
    \setlength{\tabcolsep}{0.9pt}
    \renewcommand{\arraystretch}{1.3}
    \resizebox{\linewidth}{!}{
        \setlength{\tabcolsep}{1mm}{
            \begin{tabular}{l|ccc|ccc|ccc|ccc|ccc|ccc|ccc}
                \hline
                \multirow{2}[3]{*}{Methods}                & \multicolumn{3}{c|}{Actin Filament} & \multicolumn{3}{c|}{Actom. Bundle} & \multicolumn{3}{c|}{Cell Membrane} & \multicolumn{3}{c|}{Desmosome} & \multicolumn{3}{c|}{DNA} & \multicolumn{3}{c|}{Endop. Reticulum} & \multicolumn{3}{c}{Golgi Apparatus} \bigstrut[b]                                                                                                                              \\
                \cline{2-22}                               & MSE                                 & MAE                                & $R^2$                              & MSE                            & MAE                      & $R^2$                                 & MSE                                              & MAE   & $R^2$ & MSE   & MAE   & $R^2$ & MSE   & MAE   & $R^2$ & MSE   & MAE   & $R^2$ & MSE   & MAE   & $R^2$ \bigstrut    \\
                \hline
                Multi-Net($\mathbf{r}=1$) \cite{fnet}      & .4241                               & .4716                              & .5695                              & .7247                          & .4443                    & .2606                                 & .5940                                            & .4351 & .3930 & .8393 & .5640 & .0162 & .5806 & .5033 & .3822 & .4635 & .4914 & .5262 & .8023 & .5732 & .0801 \bigstrut[t] \\
                Tgnet-3-to-2D($\mathbf{r}=2$) \cite{Tgnet} & .3896                                 & .4476                                & .6044                                & .6670                            & .4080                      & .3196                                   & .5079                                              & .4160   & .4810   & .8379   & .5646   & .0175   & .4873   & .4639   & .4808   & .4219   & .4648   & .5688   & .7822   & .5754   & .1028                \\
                Tgnet-3-to-2D($\mathbf{r}=4$)              & .3928                                 & .4492                                & .6011                                & .6726                            & .4084                      & .3138                                   & .5335                                              & .4223   & .4548   & .8385   & .5633   & .0168   & .4797   & .4587   & .4888   & .4372   & .4736   & .5532   & .7931   & .5790   & .0905                \\
                Tgnet-3-to-2D($\mathbf{r}=8$)              & .4107                                 & .4570                                & .5829                                & .6973                            & .4166                      & .2886                                   & .5529                                              & .4298   & .4350   & .8408   & .5637   & .0141   & .5087   & .4720   & .4578   & .4449   & .4748   & .5453   & .7975   & .5777   & .0853   \bigstrut[b] \\
                \hline
            \end{tabular}%
        }}
    \resizebox{\linewidth}{!}{
        \setlength{\tabcolsep}{1mm}{

            \begin{tabular}{l|ccc|ccc|ccc|ccc|ccc|ccc|ccc}
                \hline
                \multirow{2}[4]{*}{Methods}                & \multicolumn{3}{c|}{Microtubule} & \multicolumn{3}{c|}{Mitochondria} & \multicolumn{3}{c|}{Nuclear Envelope} & \multicolumn{3}{c|}{Nucleolus} & \multicolumn{3}{c|}{Tight Junction} & \multicolumn{3}{c|}{All} & \multicolumn{3}{c}{$\Delta_{\text{Imp}} \ (\%)$} \bigstrut                                                                                                                              \\
                \cline{2-22}                               & MSE                              & MAE                               & $R^2$                                 & MSE                            & MAE                                 & $R^2$                    & MSE                                                        & MAE   & $R^2$ & MSE   & MAE   & $R^2$ & MSE   & MAE   & $R^2$ & MSE   & MAE   & $R^2$ & MSE   & MAE   & $R^2$ \bigstrut    \\
                \hline
                Multi-Net($\mathbf{r}=1$) \cite{fnet}      & .3682                            & .4348                             & .6296                                 & .4684                          & .3921                               & .5172                    & .3014                                                      & .3006 & .6954 & .2164 & .1789 & .7826 & .6474 & .3369 & .3370 & .5341 & .4269 & .4337 & 0.000 & 0.000 & 0.000 \bigstrut[t] \\
                Tgnet-3-to-2D($\mathbf{r}=2$) \cite{Tgnet} & .3230                              & .4082                               & .6750                                   & .4726                            & .4058                                 & .5129                      & .2783                                                        & .2892   & .7186   & .1995   & .1679   & .7996   & .6295   & .3262   & .3554   & .4995   & .4113   & .4693   & 6.4782   & 3.6543   & 8.2084                \\
                Tgnet-3-to-2D($\mathbf{r}=4$)              & .3546                              & .4292                               & .6432                                   & .5011                            & .4132                                 & .4835                      & .3044                                                        & .2968   & .6923   & .2101   & .1708   & .7890   & .6699   & .3322   & .7926   & .5158   & .4166   & .4526   & 3.4263   & 2.4127   & 4.3579                \\
                Tgnet-3-to-2D($\mathbf{r}=8$)              & .3401                              & .4165                               & .6578                                   & .5403                            & .4249                                 & .4431                      & .3471                                                        & .3149   & .6491   & .2356   & .1825   & .7633   & .7520   & .3491   & .2301   & .5386   & .4231   & .4292   & -0.8425   & .8901   & -1.037 \bigstrut[b]   \\
                \hline
            \end{tabular}%
        }}

    \caption{
        \textbf{Comparisons of different structures on diverse sparsity ratios.}
        We utilize the FNet as the reliable baseline to metric our proposals. The Tgnet-3-to-2D is the best setting in SparseSSP, and its performance on different subcellular structures achieves a reliable level for actual biological research. In sparsity ratio 8, i.e., reducing 87.5\% imaging times, SparseSSP only shows a minor performance decay.
    }

    \label{tab:full_cells}
    % \vspace{-10mm}
    % \vspace{-5mm}
\end{table*}

\subsection{Analyzation of Resource Consumption}
\label{sec:compute}
SSP is a 3D dense regression problem, and it is logical to utilize the 3D networks to solve it. The proposed hybrid dimensions topologies incorporate dimension transformation, facilitating the replacement of 3D operations and significantly reducing computational costs (including the training and inference procedures).
We demonstrate details of resource consumption on different topologies based on the experiments under sparsity ratio 2.
In \cref{tab:cost} it can be seen that pure 2D topology is the most efficient.
However, its evaluation metrics are poorer than those of other topologies.
Hybrid dimension topologies cost less resources and achieve excellent performance compared to 3D.
The pure 3D topology costs larger MACs and memory than others. The 2-to-3D surpasses the 3-to-2D in resource efficiency due to the replacement of the feature extractor with 2D layers. Although 2-to-3D is slightly behind 3-to-2D in performance, it trades off performance and efficiency, which is the most cost-effective solution.

\subsection{Different Subcellular Structures}

% \noindent
% \textbf{Visualization.}

\subsubsection{Visualization.}
We compare the prediction results under dense-view and sparse-view input in \cref{fig:visz}.
Without staining or using SSP, it is difficult for biologists to directly observe subcellular structures from TL images, as shown in column 1.
As shown in \cref{fig:visz}, the sparse-view SSP can show the position and shape of organelles and fluorescent staining.
From the visualization results, SSP is a reliable alternative to fluorescent staining.
It can be observed that with the sparse view in higher ratios, the prediction results maintain the quality well with dense views, e.g., the Nucleoli, Microtubule, and NuclearEnvelope.
While for the slender striped structures, e.g., ActomyosinBundle, sparse-view would cause a few deficiencies.
This is primarily due to the electronic noise of the fluorescence label adversely affecting the learning of ActomyosinBundle.

\noindent
\textbf{Quantitative Comparison.}
% \subsubsection{Quantitative Comparison.}
\cref{tab:full_cells} reports the detailed results of 12 datasets.
FNet \cite{fnet} (i.e., the Multi-Net in the table) is sufficient to assist with biological research; thus, we set it as the reliable baseline for application use on different subcellular structures.
Following RepMode, we utilize a relative performance improvement metric over Multi-Net (i.e., $\Delta_{Imp}$).
We compare the best solution in hybrid dimensions topology, i.e., Tgnet-3-to-2D, on diverse sparsity ratios.
Our framework proposal (i.e., hybrid dimensions topology) demonstrates keeping the small MSE and MAE with reduced imaging frequency.

\section{Conclusion}
\label{sec:clu}
We propose a novel framework, SparseSSP, featuring hybrid dimensions topology to achieve fast and efficient SSP using sparse transmitted light images, thus reducing the need for extensive microscopy imaging and computational costs.
SparseSSP leverages a one-to-many voxel mapping paradigm to implicitly refine sparse Z-axis information.
Unlike prior works that rely solely on pure 3D topologies, our approach integrates hybrid 3D and 2D network layers to enhance efficiency, striking a balance between computational cost and prediction quality. 
By incorporating dimension transformation into the network, we convert Z-axis information into channel features, allowing advanced 2D modules to effectively address 3D SSP challenges.
Furthermore, we explore the potential of migrating 2D backbones to SparseSSP.
Extensive experiments with SOTA multi-task methods across various sparse ratios demonstrate the effectiveness and advantages of our approach. 
To the best of our knowledge, this work is pioneering in utilizing sparse-view TL images for SSP, significantly advancing the visualization of rapid biological dynamics on low-cost devices and samples.

\bibliographystyle{splncs04}
\bibliography{main}
\end{document}